\documentclass[twocolumn]{bmcart}

\usepackage[utf8]{inputenc} 

\usepackage{algorithm}
\usepackage{xcolor}
\usepackage{algpseudocode}
\usepackage{xspace, multirow}
\usepackage{amsmath, soul}
\usepackage{amsfonts}
\usepackage{./style/optidef}
\usepackage{graphicx}
\usepackage[flushleft]{threeparttable}
\usepackage{comment}
\usepackage{tikz}
\usepackage{ url}

\graphicspath{
{./figure/}
{./figure/temp/}
{./figure/figure_pdf/}
{./figure/K_cosim_D2k/Order_A_D/}
{./figure/K_cosim_D2k/Order_B_C/}
{./figure/K_cosim_D2k/Freq_A_D/}
{./figure/K_cosim_D2k/OR_B_C/}
{./figure/K_cosim_D4k/Order_A_D/}
{./figure/K_cosim_D4k/Order_B_C/}
{./figure/K_cosim_D4k/Freq_A_D/}
{./figure/K_cosim_D4k/OR_B_C/}
{./figure/K_cosim_D6k/Order_A_D/}
{./figure/K_cosim_D6k/Order_B_C/}
{./figure/K_cosim_D6k/Freq_A_D/}
{./figure/K_cosim_D6k/OR_B_C/}
 }
\startlocaldefs

\newcommand{\drug}{\mbox{$\mathop{\mathtt{d}}\limits$}\xspace}
\newcommand{\drugi}{\mbox{$\mathop{\mathtt{d}_i}\limits$}\xspace}
\newcommand{\drugj}{\mbox{$\mathop{\mathtt{d}_j}\limits$}\xspace}
\newcommand{\D}{\mbox{$\mathop{D}\limits$}\xspace}
\newcommand{\Dp}{\mbox{$\mathop{D_p}\limits$}\xspace}
\newcommand{\DpII}{\mbox{$\mathop{D^{(2)}_p}\limits$}\xspace}
\newcommand{\DpN}{\mbox{$\mathop{D^{(n)}_p}\limits$}\xspace}
\newcommand{\Gp}{\mbox{$\mathop{\mathcal{G}_p}\limits$}\xspace}
\newcommand{\Dq}{\mbox{$\mathop{D_q}\limits$}\xspace}
\newcommand{\DqII}{\mbox{$\mathop{D^{(2)}_q}\limits$}\xspace}
\newcommand{\DqN}{\mbox{$\mathop{D^{(n)}_q}\limits$}\xspace}
\newcommand{\Gq}{\mbox{$\mathop{\mathcal{G}_q}\limits$}\xspace}

\newcommand{\OR}{\mbox{$\mathop{\mathtt{OR}}\limits$}\xspace}
\newcommand{\ORs}{\mbox{$\mathop{\mathtt{OR}}\limits$s}\xspace}

\newcommand{\cmi}{\mbox{$\mathop{\boldsymbol{c}_i}\limits$}\xspace}
\newcommand{\cmj}{\mbox{$\mathop{\boldsymbol{c}_j}\limits$}\xspace}
\newcommand{\cmADRi}{\mbox{$\mathop{\boldsymbol{c}^{+}_i}\limits$}\xspace}
\newcommand{\cmNoADRi}{\mbox{$\mathop{\boldsymbol{c}^{-}_i}\limits$}\xspace}
\newcommand{\adrs}{\mbox{$\mathop{\text{ADRs}}\limits$}\xspace}
\newcommand{\adr}{\mbox{$\mathop{\text{ADR}}\limits$}\xspace}

\newcommand{\ddis}{\mbox{$\mathop{\text{DDIs}}\limits$}\xspace}
\newcommand{\ddi}{\mbox{$\mathop{\text{DDI}}\limits$}\xspace}

\newcommand{\SDSd}{\mbox{$\mathop{\text{SDS}_{\text{2d}}}\limits$}\xspace}
\newcommand{\SDSc}{\mbox{$\mathop{\text{SDS}_{\text{cm}}}\limits$}\xspace}
\newcommand{\SDS}{\mbox{$\mathop{\text{SDS}}\limits$}\xspace}
\newcommand{\SDSs}{\mbox{$\mathop{\text{SDSs}}\limits$}\xspace}

\newcommand{\Sgm}{\mbox{$\mathop{\mathcal{S}_\text{gm}}\limits$}\xspace}
\newcommand{\GMop}{\mbox{$\mathop{\mathcal{K}_\text{gm}}\limits$}\xspace}
\newcommand{\GMopcm}{\mbox{$\mathop{\mathcal{K}_\text{gm}^\text{cm}}\limits$}\xspace}
\newcommand{\GMavg}{\mbox{$\mathop{\mathcal{K}_\text{ds}}\limits$}\xspace}
\newcommand{\GMprob}{\mbox{$\mathop{\mathcal{K}_\text{pb}}\limits$}\xspace}
\newcommand{\GMtan}{\mbox{$\mathop{\mathcal{K}_\text{cd}}\limits$}\xspace}

\newcommand{\GMtanorderI}{\mbox{$\mathop{\mathcal{K}^{(1)}_\text{cd}}\limits$}\xspace}
\newcommand{\GMtanorderII}{\mbox{$\mathop{\mathcal{K}^{(2)}_\text{cd}}\limits$}\xspace}

\newcommand{\Dall}{\mbox{$\mathop{\mathcal{D}_{\text{FAERS}}}\limits$}\xspace} 

\newcommand{\Dataset}{\mbox{$\mathop{\mathcal{D}^{*}}\limits$}\xspace}
\newcommand{\Dscider}{\mbox{$\mathop{\mathcal{D}_{\text{Myo}}}\limits$}\xspace}

\newcommand{\IID}{\mbox{$\mathop{\text{2d}}\limits$}\xspace}
\newcommand{\coSim}{\mbox{$\mathop{\text{cm}}\limits$}\xspace}
\newcommand{\orderI}{\mbox{$\mathop{\text{ord-}1}\limits$}\xspace}
\newcommand{\orderII}{\mbox{$\mathop{\text{ord-}2}\limits$}\xspace}

\newcommand{\Nneg}{\mbox{$\mathop{\mathcal{N}}\limits$}\xspace}  
\newcommand{\Nminus}{\mbox{$\mathop{\mathcal{N}^-}\limits$}\xspace}  
\newcommand{\Nzero}{\mbox{$\mathop{\mathcal{N}^0}\limits$}\xspace}  
\newcommand{\Mpos}{\mbox{$\mathop{\mathcal{M}}\limits$}\xspace}  
\newcommand{\Mzero}{\mbox{$\mathop{\mathcal{M}^0}\limits$}\xspace}  
\newcommand{\Mplus}{\mbox{$\mathop{\mathcal{M}^+}\limits$}\xspace}  
\newcommand{\Nmis}{\mbox{$\mathop{\tilde{\mathcal{N}}^{+}}\limits$}\xspace}
\newcommand{\Mmis}{\mbox{$\mathop{\tilde{\mathcal{M}}^{-}}\limits$}\xspace}  
\newcommand{\Nmist}{\mbox{$\mathop{\tilde{\mathcal{N}}^{10+}}\limits$}\xspace}
\newcommand{\Mmist}{\mbox{$\mathop{\tilde{\mathcal{M}}^{10-}}\limits$}\xspace}   
\newcommand{\Nt}{\mbox{$\mathop{\tilde{\mathcal{N}}^{10-}}\limits$}\xspace}
\newcommand{\Mt}{\mbox{$\mathop{\tilde{\mathcal{M}}^{10+}}\limits$}\xspace} 
\newcommand{\Asix}{\Nminus}  
\newcommand{\Bsix}{\Nzero}  
\newcommand{\Csix}{\Mzero}  
\newcommand{\Dsix}{\Mplus}  

\newcommand{\mytitle}{Drug-drug interaction prediction based on \\co-medication patterns and graph matching}
\endlocaldefs

\begin{document}

\begin{frontmatter}

\begin{fmbox}
\dochead{Research}


\title{\mytitle}


\author[
   addressref={aff1},                   
   email={chiangwe@iupui.edu}   
]{\inits{WC}\fnm{Wen-Hao} \snm{Chiang}}
\author[
   addressref={aff2},
   email={Li.Shen@pennmedicine.upenn.edu}
]{\inits{LS}\fnm{Li} \snm{Shen}}
\author[
   addressref={aff3},
   email={Lang.Li@osumc.edu}
]{\inits{LS}\fnm{Lang} \snm{Li}}
\author[
   addressref={aff4},
   corref={aff4},
   email={xning@iupui.edu}
]{\inits{XN}\fnm{Xia} \snm{Ning}}


\address[id=aff1]{
  \orgname{Department of Computer \& Information Science, Indiana University - Purdue University Indianapolis}, 
  \postcode{46202}                                
  \city{Indianapolis},                              
  \cny{USA. Email: chiangwe@iupui.edu}                                    
}
\address[id=aff2]{%
  \orgname{Department of Biostatistics, Epidemiology and Informatics, University of Pennsylvania},
  \postcode{19104}
  \city{Philadelphia},
  \cny{USA. Email: Li.Shen@pennmedicine.upenn.edu}
}
\address[id=aff3]{%
  \orgname{Department of Biomedical Informatics, Ohio State University},
  \postcode{43210}
  \city{Columbus},
  \cny{USA. Email: Lang.Li@osumc.edu}
}
\address[id=aff4]{
  \orgname{Department of Computer \& Information Science, Indiana University - Purdue University Indianapolis}, 
  \postcode{46202}                                
  \city{Indianapolis},                              
  \cny{USA. Email: xning@iupui.edu}                                    
}


\begin{artnotes}
\end{artnotes}



\begin{abstractbox}
\begin{abstract} 
\parttitle{Background}
The problem of predicting whether a drug combination of arbitrary orders is
likely to induce adverse drug reactions is considered in this manuscript.
\parttitle{Methods}
Novel kernels over drug combinations of arbitrary orders are developed 
within support vector machines for the prediction.
Graph matching methods
are used in the novel kernels to measure the similarities among drug combinations, in which
drug co-medication patterns are leveraged to measure single drug similarities.
\parttitle{Results}
The experimental results on a real-world dataset demonstrated that the new kernels achieve
an area under the curve (AUC) value 0.912 for the prediction problem. 
\parttitle{Conclusions}
The new methods with drug co-medication based single drug similarities can accurately predict whether a drug combination is likely to 
induce adverse drug reactions of interest.
%
%
%
\end{abstract}



\end{abstractbox}
\end{fmbox}

\end{frontmatter}



\section*{Introduction}
\label{sec:intro}
%
Drug-Drug Interactions (\ddis) and the associated Adverse Drug Reactions (ADRs) represent 
a consistent detriment to the public health in the United States. 
\ddis have accounted for approximately 26\% of the ADRs,
occurred among 50\% of the hospitalized patients~\cite{Ramirez2010}, 
and caused nearly 74,000 emergency room visits and 195,000 hospitalizations annually in the US~\cite{Percha2013}. 
Apart from these, because of the common practice of co-medication among 
elderly Americans, particularly co-medication of more than two drugs, the high-order
drug-drug interactions and their associated ADRs have imposed significant scientific and
public health challenges. 
The National Health and Nutrition Examination Survey~\cite{Survey}
reports that more than 76\% of the 
elderly Americans take two or more drugs every day. 
Another study~\cite{Iyer2014} estimates that 
about 29.4\% of elderly American patients take six or more drugs every day. 
However, for most of such high-order \ddis, their mechanisms are unknown.

%
In this manuscript, novel approaches to predicting
whether high-order drug combinations are likely to induce ADRs are presented.
The prediction problems are formulated as a binary classification problem
and support vector machines (SVMs) are used for the prediction.
%
Novel kernels over drug combinations of arbitrary orders are developed within
the framework of SVMs.
These kernels are constructed using drug
co-medication information to measure single drug similarities
and graph matching on drug combination graphs to measure drug combination similarities.
A comparison on the new kernels with other convolutional kernels
and probabilistic kernels on drug combinations is also conducted.
%
The experimental results demonstrate that the new kernels outperform the others and 
can accurately predict whether
a drug combination is likely to induce ADRs of interest with an AUC value 0.912. 
To the best of our knowledge, this manuscript represents the first effort in predicting
\ddis for drug combinations of arbitrary orders.

\section*{Background}
\label{sec:work}
%

\subsection*{Drug-drug interactions}

Significant research efforts have been dedicated to detect pairwise
drug-drug interactions (\ddis)~\cite{Vilar2014,Hammann2014} in recent years.
%
Existing methods either extract \ddi pairs mentioned
in medical literature or Electronic Health Records (EHRs)~\cite{Iyer2014}, 
or predict/score \ddi pairs from various drug/target information~\cite{Luo2014}.
%
While most of the existing \ddi studies are focused on interactions between a pair of drugs
(i.e., order-2 \ddis), 
%
understanding high-order \ddis and their associated ADRs has attracted increasing attention 
recently~\cite{Percha2013, Harpaz2012}. 
These emerging methods on high-order \ddi studies 
are largely focused on how to discover high-order \ddis through mining frequent itemsets
(i.e., drug combinations) from EHRs efficiently. 
Most recent work also includes pattern discovery from directional high-order
\ddis~\cite{Ning2017a} and directional high-order \ddi prediction~\cite{Ning2017b}.

\subsection*{Graph matching}

%
  Graph matching is to find the optimal vertex correspondence between two
  graphs~\cite{Conte2004, Forggia2014}. 
  Graph matching problems can be broadly classified into two categories. 
  The first category is exact graph matching, which is to find the graph and subgraph
  isomorphisms so that the mapping of vertices between two graphs is bijective and edge-preserving (i.e.,
  vertices connected by an edge in one graph are mapped to vertices in the other graph that are
  also connected by an edge).
  The second category is inexact graph matching, which allows errors (e.g., different types of matched
  vertices in attributed graphs) during matching, and thus it is to minimize the total errors in
  finding optimal graph matching.
  Typical algorithms for graph matching include spectral methods~\cite{Caelli2004}, probabilistic
  methods~\cite{Caetano2004}, tree search~\cite{Messmer1998}, etc. 
\section*{Definitions and notations}
\label{sec:def}

We use $d_i$ to represent a drug,
and \mbox{$D^k = \{d_1, d_2, \cdots, d_k\}$} to represent a combination of $k$ drugs,
where $k$ is the number of unique drugs in $D^k$ (i.e., $k = |D^k|$)
and thus the order of $D^k$.
A drug combination $D^k$ is defined when the drugs and only the drugs in $D^k$ are taken simultaneously.
There are no orderings among the drugs in a drug combination.
When no ambiguity is raised, we drop the superscript $k$ in $D^k$ and represent a drug combination as
$D$.
An event is referred to as a patient taking a drug combination.
In addition, in this manuscript, 
all vectors (e.g., $\mathbf{c}$) are represented
by bold lower-case letters and all matrices (e.g., $X$) are represented
by upper-case letters. Row vectors are represented
by having the transpose superscript $^{\mathsf{T}}$, otherwise by default they
are column vectors.
Table~\ref{tbl:Notation} summarizes the important notations in the manuscript.

%

\section*{Methods}
\label{sec:methods}

We formulate the problem of predicting whether high-order drug combinations induce a particular
\adr as a binary classification problem, and solve the classification problem within the framework
of kernel methods and support vector machines (SVMs). 
In this manuscript, we consider myopathy as the \adr in particular.
The central concept of SVM-based classification methods is that ``similar'' instances
are likely to share similar labels, and thus the key is to capture and measure the ``similarities''
among instances (i.e., drug combinations in our \adr prediction problem) via kernels.
In the case of drug combinations, we hypothesize that if two drug combinations share similar
pharmaceutical, pharmacokinetic and/or pharmacodynamic properties, they may induce similar \adrs.
Therefore, the question boils down to effectively representing and measuring the similarities
in terms of such properties.
To this end, we develop various kernels over drug combinations. A key property of such kernels as will
be discussed later is that they are able to deal with drug combinations of arbitrary orders.
These kernels are constructed using single drug similarities, which
incorporate various drug information that could relate to \ddis. 
Here we decompose the discussion on such kernels from three aspects:
1). single drug similarities (\SDS) as in Section~\ref{sec:methods:sds},
2). our new kernel based on matching similar drugs in drug combination graphs
in Section~\ref{sec:methods:gmkernel}, and
3). other convolutional kernels~\cite{Haussler1999} in Section~\ref{sec:methods:otherkernel}. 
Given these kernels, we further employ the freely available SVM-\textit{Light} software to build up 
the binary classifiers and conduct our experiments based on such classifiers~\cite{Joachims1998}.

\subsection*{Single drug similarities}
\label{sec:methods:sds}

We use two different approaches to measuring single drug similarities (\SDS). The first approach
measures single drug similarities based on their intrinsic properties that can be represented
by their 2D structures~\cite{Wale2008}. The second approach measures the similarities in a more
data-driven fashion based on the co-occurrence patterns among drugs.

%
%




\subsubsection*{\SDS from drug 2d structures}
\label{sec:methods:sds:2d}

A straightforward way to measure \SDSs between two drugs is to look at their structures, which
ultimately determine their physicochemical properties.
We use Extended Connectivity Fingerprints (ECFP)~\cite{Scitegic}
of length 2,048 to represent drug 2D structures.
Each of the fingerprint dimensions corresponds to a substructure among the drugs of interest.
The binary values in the fingerprints represent whether a drug has the corresponding substructure
or not. We use a vector $\boldsymbol{x}_i \in \mathbb{R}^{2048}$ to represent the fingerprint
for drug \drugi.
The \SDS between two drugs from their 2D structures, denoted as \SDSd,
is calculated as the Tanimoto coefficient between
their ECFP fingerprints~\cite{Willett98}.
Tanimoto coefficient between two sets is defined as follows,
\begin{equation}
  \label{eq:tanimoto}
  \text{Tanimoto}(S_1, S_2) =
  \frac{|S_1\cap S_2|}{|S_1|+|S_2|-|S_1\cap S_2|},
\end{equation}
where $|S|$ is the cardinality of set $S$. Thus, \SDSd is defined as
\begin{eqnarray}
  \begin{aligned}
    \label{eq:sds2d}
    \SDSd(\drug_i,\drug_j) =
    \text{Tanimoto}(\{\boldsymbol{x}_i\}, \{\boldsymbol{x}_j\}),
  \end{aligned}
\end{eqnarray}
where $\{\boldsymbol{x}_i\}$ represents the set of substructures that $\drug_i$ has in
its fingerprint $\boldsymbol{x}_i$.

\subsubsection*{\SDS based on co-medications}
\label{sec:methods:sds:cooccur}  
%

We develop a new approach to measuring the \SDS between two drugs by looking at whether they are
often involved in co-medications with similar other drugs, respectively.
The hypothesis is that drugs that are respectively taken together with other similar drugs
may share similar therapeutic purposes and target similar therapeutic targets, and thus
behave similarly in inducing \adrs.
Such data-driven co-medication based \SDSs have a potential advantage over \SDSd in that they
leverage the signals from \adrs information directly that may not be captured or explained
by drug 2D structures or other features on individual drugs.
Such co-medication based \SDS is denoted as \SDSc.

We use two vectors $\cmADRi \in \mathbb{R}^n$ and $\cmNoADRi \in \mathbb{R}^n$ ($n$ is the
total number of drugs) to represent the co-medication information for drug \drugi.
The $j$-th dimension ($j = 1, \cdots, n$) in \cmADRi/\cmNoADRi corresponds to drug \drugj, 
and the value on the $j$-th dimension in \cmADRi/\cmNoADRi is the co-medication frequency of
\drugi and \drugj in all the events with/without \adrs. Both \cmADRi and \cmNoADRi values
are then normalized into probabilities. The normalized \cmADRi and \cmNoADRi are further
concatenated into one vector \cmi, that is, $\cmi = [\cmADRi; \cmNoADRi]$, for \drugi.
The \SDSc between drug \drugi and \drugj is calculated as the cosine similarity between
\cmi and \cmj.
The reason why we use \cmADRi and \cmNoADRi to construct \cmi instead of co-medication
frequencies from all events with and without \adrs together is that the co-medication patterns
from the two types of events can be very different, and thus one unified co-medication
vector for both of them could not necessarily capture discriminative information among
drugs.

\subsection*{Drug combination kernels from graph matching}
\label{sec:methods:gmkernel}
%
We formulate the problem of comparing drug combination similarities through matching
drug combination graphs, and develop a graph-matching based kernel for drug combination
similarities.
Specifically, for a drug combination
\mbox{$\Dp = \{\drug_{p1}, \drug_{p2}, \cdots, \drug_{pk_p}\}$},
we construct a complete graph \Gp of $k_p$ nodes,
in which each node represents a drug in \Dp, and all the nodes are
connected to one another. Thus, the similarity between drug combination \Dp and \Dq can be measured
based on how \Gp and \Gq match to each other.
In matching such graphs, we consider \SDSs so that drugs that
are similar to each other should be matched, and the graph matching procedure should maximize
the overall \SDSs from matched drugs.
The underlying assumption is that if two drug
combinations share similar drugs, they could have similar \adrs.
Figure~\ref{fig:gmatch} illustrates the idea of complete graph matching for two drug combinations,
in which the drugs connected by dash lines are matched between \Dp and \Dq.
The similarity calculated from graph matching over two drug combinations, denoted as \Sgm,
will the sum of \SDSs from matched drugs. 
\Sgm will be further converted to a valid kernel, denoted as \GMop.
%

\subsubsection*{Graph matching algorithm for \GMop}
\label{sec:methods:gmkernel:algo}

The drug combination graph matching problem 
can be solved as a well known linear sum assignment problem (LSAP)~\cite{Burkard1980}.
The objective is to minimize the total cost of matching vertices in two graphs, and thus to
find the graph matching with minimal total cost.
In the case of high-order drug combinations, we define the cost of matching two drugs
\drugi and \drugj as the dis-similarity between the drugs, that is,
\begin{equation}
  \label{eq:cost}
  cost(\drugi, \drugj) = 1 - \SDS(\drugi, \drugj),
\end{equation}
where $cost(\drugi, \drugj)$ is the cost between \drugi and \drugj, 
\SDS can be either \SDSd or \SDSc.
Thus, if two drugs are very similar (i.e., large \SDS),
the cost of matching them will be small and therefore they are more likely to be matched.

Therefore, the graph matching can be solved by solving the following LSAP problem:
\begin{equation}
  \label{opt:lsap}
  \begin{aligned}
    & \underset{{X}}{\min} & & \text{trace}(C(\Gp, \Gq){X}^{\mathsf{T}} ) \\
    & \text{subject to}    & & {X}\in\mathcal{P}, \\
    &                      & &
    \mathcal{P}\coloneqq}{\{{X}\mid {X}\in \mathbb{R}^{k_p\times k_q}, {X}_{i, j}\in\{0,1\}, \\
    &                      & & \quad\quad\quad\quad\sum_{i=1}^{k_p} {X}_{i,j} \le 1,\sum_{j=1}^{k_q} {X}_{i,j} \le 1, \\
    &                      & & \quad\quad\quad\quad\sum_{i=1}^{k_p} \sum_{j=1}^{k_q} {X}_{i,j} = \min(k_p, k_q)\},
  \end{aligned}
\end{equation}
where $\text{trace}()$ is the trace of a matrix; 
and $k_p$ and $k_q$ are the number of vertices in \Gp and \Gq (and thus
the order of \Dp and \Dq), respectively;  
\mbox{$C(\Gp, \Gq) \in \mathbb{R}^{k_p \times k_q}$} is the pairwise drug-matching cost matrix for
two drug combinations \Dp and \Dq (\mbox{$C(i, j) = cost(\drug_{pi}, \drug_{qj})$}, $\drug_{pi} \in \Dp$,
$\drug_{qj} \in \Dq$).
In Problem~\ref{opt:lsap}, $X$ is the assignment matrix to match \Gp and \Gq (i.e., to assign a vertex
in \Gp to a vertex in \Gq),
in which all the values are either 0 or 1, both the row sum and the column sum are either 0 or 1
(i.e., a vertex is either matched or not; if it is matched, it is matched to only one
vertex in the other graph), and thus the sum of all the values is exactly the minimal of $k_p$ and $k_q$
(i.e., the vertices in the small graph have to be all matched).
Essentially, $X$ assigns each of the vertices in the smaller graph of \Gp and \Gq to exactly one vertex
in the larger graph.
The optimization problem in~\ref{opt:lsap} can be solved by the Hungarian algorithm~\cite{Kuhn1955}.
The drug-combination similarity \Sgm is then calculated as
\begin{equation}
  \label{eq:sgm}
  \Sgm(\Dp, \Dq) = \text{trace}( J - C(\Gp, \Gq){X}^{\mathsf{T}}),
\end{equation}
where ${J}\in \mathbb{R}^{k_p\times k_q}$ is a matrix of all 1's.

The drug-combination similarity matrix \Sgm is always symmetric but not necessarily positive
semi-definite, and thus not always a valid kernel. To convert \Sgm to a valid kernel \GMop,
we follow the approach in Saigo \etal \cite{Saigo2004}.
Specifically, we first conduct an eigenvalue decomposition on \Sgm,
subtract from the diagonal of the eigenvalue matrix its smallest negative eigenvalue, and
reconstruct the original matrix from the altered decomposition.
The resulted matrix is positive, semi-definite, and is used as \GMop.

\subsection*{Convolutional drug-combination kernels}
\label{sec:methods:otherkernel}

\subsubsection*{Drug combination kernels from common drugs}
\label{sec:methods:tanimoto}
%
We define a drug-combination kernel, denoted as \GMtan, based on common drugs among drug combinations.
\GMtan is calculated as the Tanimoto coefficient over the sets of drugs in the drug combinations,
that is,
\begin{equation}
  \label{eq:gmtan}
  \GMtan(\Dp, \Dq) = \text{Tanimoto}(\Dp, \Dq),
\end{equation}
where $\text{Tanimoto}()$ is defined as in Equation~\ref{eq:tanimoto}. It has been proved that
Tanimoto coefficient is a valid kernel function~\cite{Pizzuti2009}.
\GMtan essentially measures the proportion of shared common drugs among two drug combinations. The
underlying assumption is that if two drug combinations share many common drugs, they are likely to
have similar properties.
To further enhance the similarity between two drug combinations from their common drugs, we
also define an order-2 \GMtan of drug combinations, denoted as \GMtanorderII
(\GMtan in Equation~\ref{eq:gmtan} is correspondingly referred to as order-1 \GMtan and denoted as
\GMtanorderI).
We first represent a drug combination $\D = \{\drug_1, \drug_2, \cdots, \drug_k\}$ by all its single
drugs and drug pairs, denoted as
\mbox{$\D^{(2)} = \{\drug_1, \drug_2, \cdots, \drug_k, (\drug_1, \drug_2), (\drug_1, \drug_3), \cdots,(\drug_{k-1}, \drug_k)\}$}.
Thus, \GMtanorderII on two drug combinations \Dp and \Dq can be calculated as the
Tanimoto coefficient on $\DpII$ and $\DqII$, that is,
\begin{equation}
  \label{eq:gmtan2}
  \GMtanorderII(\Dp, \Dq) = \text{Tanimoto}(\DpII, \DqII).
\end{equation}
Intuitively, \GMtanorderII better differentiates drug combinations with many shared drugs
from those with fewer shared drugs than \GMtanorderI.
We only extend \GMtan to order 2 since higher-order extension does not lead to 
better performance according to our experimental results.
According to Equation~\ref{eq:gmtan}, when the order becomes much higher, 
\text{Tanimoto}(\DpN, \DqN) may become very small due to a rapid combinatorial
growth in the denominator and 
the insufficient common drug $n$-tuples (i.e., the number in the nominator).
Thus, \GMtan with extension to much higher order may lose the ability to
differentiate drug combinations that contain more common drugs.

\subsubsection*{Drug combination kernels from drug similarities}
\label{sec:methods:svgsds}

The drug combination similarities can also be measured by the average drug similarities.
The hypothesis is that if two drug combinations have drugs that are similar on average, 
they may share similar properties.
If two drug combinations have drugs that are similar on average, they
may share similar properties.
Therefore, we define an average-drug-similarity based kernel for drug combinations, denoted as \GMavg,
as follows,
\begin{equation}
  \label{eq:avgsds}
  \GMavg(\Dp, \Dq) = \frac{1}{k_pk_q} \sum_{\scriptsize{\drug_i \in \Dp}}\sum_{\scriptsize{\drug_j \in \Dq}} \SDS(\drug_i, \drug_j),
\end{equation}
where $k_p$ and $k_q$ are the order of \Dp and \Dq, respectively, and \SDS can be \SDSd or \SDSc.
Intuitively, \GMavg  tends to capture averaged and smoothed
drug combination similarities.
It has been proved that as long as the involved \SDSs are valid kernels
(i.e., positive semi-definite),
\GMavg  will also be a valid kernel \cite{Haussler1999}.

\subsubsection*{Probabilistic drug combination kernels from drug sets}
\label{sec:methods:sets}

We apply an ensemble kernel for drug combinations based on the idea as in~\cite{Zhou2006}.
The key idea is to use a reproducing kernel to characterize sample similarities (i.e., \SDS),
and to use a probabilistic distance in the reproducing kernel Hilbert space (RKHS)
to measure the ensemble similarity. The resulted ensemble similarity matrix is a valid kernel
matrix, denoted as \GMprob. 
This ensemble involves an eigen value decomposition, during which,
it is possible that some similarity matrices are deprecated numerically and it leads
to defeats in \GMprob calculation.
To deal with this issue, we increase the diagonals of involved square matrices by a small value
to guarantee the positive semi-definite properties.
%

\section*{Materials}
\label{sec:data}

\subsection*{Mining drug combinations}
\label{sec:data:mining}

We extract high-order drug combinations from FDA Adverse Event Reporting System (FAERS)~\cite{FAERS}. 
%
We use myopathy as the \adr of particular interest, and extract
64,892 case (myopathy) events, in which patients report myopathy after taking multiple drugs,
and 1,475,840 control (non-myopathy) events, in which patients do not report myopathy after
taking drugs.
Each of these events involves a combination of more than one drug.
%

%

%
Among all the involved drug combinations, 10,250 unique drug combinations appear in both case and
control events. For those 10,250 drug combinations, we use Odds Ratio (\OR) to quantify their \adr risks.
The \OR for a drug combination \D is defined based on the contingency table~\ref{table:contigen},
that is, it is the ratio of the following two values:
1). the odds that the \adr occurs when \D is taken (i.e., $\frac{n_1}{m_1}$ in Table~\ref{table:contigen}); and
2). the odds that the \adr occurs when \D is not taken (i.e., $\frac{n_2}{m_2}$ in Table~\ref{table:contigen}).
\mbox{\OR $<$ 1} indicates the decreased risk of \adr after a patient takes the drug combination,
\mbox{\OR$=1$} indicates no risk change, and \mbox{\OR $>$ 1} indicates the increased risk.
In the 10,250 drug combinations, 8,986 combinations have $\OR > 1$ and 1,264 combinations have
$\OR < 1$. 
These two sets of drug combinations are denoted as \Mzero and \Nzero, respectively.
In addition to these combinations, there are 27,387 unique drug combinations
that only appear in case events and 621,449 unique drug combinations that only appear in control events.
These two sets are denoted as \Mplus and \Nminus, respectively.
The set of drug combinations in case events is denoted as \Mpos
(i.e., $\Mpos = \Mplus \cup \Mzero$), and the set of drug combinations in control events is
denoted as\Nneg (i.e., $\Nneg = \Nminus \cup \Nzero$). 
All these four sets together define a high-order drug combination dataset from FAERS, denoted as \Dall.
Table~\ref{table:Statis_whole} presents the statistics of \Dall.
%


\subsection*{Training data generation}
\label{sec:data:train}

As shown in Table~\ref{table:Statis_whole}, \Mplus and \Mzero of \Dall
have fewer drug combinations than \Nminus and \Nzero, and the drug combinations in \Mplus
are very infrequent (average frequency 1.402).
To use more frequent and more confident drug combinations from case events, we further pruned drug
combinations from \Mplus and \Mzero as follows.
From \Mplus, we retained the top 1,000 most frequent drug combinations.
For \Mzero, we applied right-tailed Fisher's exact test 
on the drug combinations
to further test the significance of their \ORs at 5\% significance level.
Then we retained drug combinations with statistically significant \ORs.
Thus, the pruned \Mplus and \Mzero contain statistically confident drug combinations,
which are very likely to induce myopathy, and therefore, these drug combinations are labeled as
positive instances for classification model learning.

We retained all 1,264 the drug combinations in \Nzero because this set is not large and contains
informative drug combinations that may or may not induce myopathy.
We further prune \Nminus and retain the top most frequent drug combinations. The drug combinations from
\Nzero and the pruned \Nminus are labeled as negative instances. To make the positive and negative
training sets balance, we retained 2,200 drug combinations from \Nminus.
The pruned dataset from \Dall is denoted as \Dataset. Table~\ref{table:Statis_whole} presents the
description of \Dataset.
\Dataset is the set of labeled drug combinations that are used for model learning.
In \Dataset, there are in total 1,210 drugs involved. 71 out of these 1,210 drugs induce myopathy
on their own based on the Side Effect Resource (SIDER)~\cite{SIDER}. 
This set of 71 drugs is denoted as \Dscider.

\subsection*{Evaluation protocol and metrics}
\label{sec:exp:eval:protocol}

The performance of the different methods is evaluated through five-fold cross validation.
The dataset is randomly split into five folds of equal size
(i.e., same number of drug combinations).
Four folds are used for model training and the rest fold is used for testing.
This process is performed five times, with one fold for testing each time.
The final result is the average out of the five experiments.


We use accuracy, precision, recall, F1 and AUC to evaluate the performance of the methods.
Accuracy is defined as the fraction of all correctly classified
instances (i.e., true positives and true negatives) over all the instances in the testing set.
Precision is defined as the fraction of correctly classified positive instances (i.e., true positives)
over all instances that are classified as positive instances (i.e., true positives and false positives).
Recall is the fraction of correctly classified positive instances (i.e., true positives)
over all positive instances in the testing set (i.e., true positives and false negatives).
F1 is the harmonic mean of precision and recall.
AUC score is the normalized area under the curve that plots the true
positives against the false positives for different thresholds
for classification~\cite{Fawcett2006}.
Larger accuracy, precision, recall, F1 and AUC values indicate better classification performance.


\section*{Results}
\label{sec:exp:results}

\subsection*{Overall performance}
\label{sec:exp:results:overall}

%
Table~\ref{table:result_6k} presents the performance comparison among the four different
kernels in combination with different single drug similarities on dataset \Dataset.  
Kernel \GMop with \SDSc outperforms others in three (i.e., accuracy, F1 and AUC)
out of five evaluation metrics. 
Specifically, in accuracy, \GMop with \SDSc outperforms the second best kernel \GMop with
\SDSd at 0.84\%. 
In F1, \GMop with \SDSc outperforms the second best kernel \GMop with \SDSd
and \GMavg with \SDSc at 0.98\%. In AUC, \GMop with \SDSc outperforms the second best kernel
order-2 \GMtan at 0.33\%.
In precision and recall, \GMop with \SDSc is the second best kernel, whereas \GMavg with \SDSc and \GMavg with \SDSd, respectively, is the best one.
Overall, \GMop with \SDSc has the best performance compared to other kernels. This indicates
that it is effective to classify drug combinations by representing and comparing them as graphs
(i.e., a set of drugs and their co-medication relation within the set),
and measuring such graph similarities using their optimal matching (i.e., the optimal
correspondence among drugs). 
In the following discussion, we use \GMopcm to represent \GMop with \SDSc. 
More experimental results on other datasets are available in the supplementary 
materials (see Additional file 1).

\subsection*{\SDS performance}

Table~\ref{table:result_6k} shows that \SDSc on average outperforms \SDSd across different
kernels (with a few exceptions on in precision for \GMavg and \GMprob).
\SDSd considers drug intrinsic 2D structures. However, drug efficacy and side effects are the
results of many complicated interactions and processes among drugs and various bioentities,
which may not be sufficiently explained only by drug 2D structures.
Compared to \SDSd, \SDSc measures drug similarity based on their co-medication patterns,
which could be regarded as a high-level abstraction and representation of drug therapeutic
properties that may or may not be explicitly explained by each drug and its intrinsic properties
independently.

In \GMtan, order-2 representation (i.e., in \GMtanorderII)
for drug combinations outperforms order-1 representation (i.e., in \GMtanorderI). In
order-2 representation, in addition to single drugs, drug pairs are also used as a feature for
a drug combination, which stresses the signals in drug combinations. This also conforms to
common observations in other applications~\cite{Min2014}, in which higher-order features
improve classification performance.

\subsection*{Classification}

Figure~\ref{fig:order_pred_AD} and~\ref{fig:order_pred_BC} present the \GMopcm
prediction values with respect to drug combination orders.
In Figure~\ref{fig:order_pred_AD}, \Mplus drug combinations have higher orders (on average
7.615 as in Table~\ref{table:Statis_whole}), and higher and mostly positive prediction values, 
while \Nminus drug combinations have lower orders (on average 2.678), and lower and mostly
negative prediction values.
Meanwhile, the mis-classification typically happens on \Nminus drug combinations of higher
orders, and on \Mplus drug combinations of lower orders. 
Similar trends apply for \Mzero and \Nzero in Figure~\ref{fig:order_pred_BC}.
This indicates that \GMop and \SDSc together are able to learn and make predictions that correspond 
to drug combination orders. In addition, drug combination order is correlated with their \adr
labels. 

Figure~\ref{fig:freq_pred_AD} presents the \GMopcm prediction values with respect
to drug combination frequencies for \Mplus and \Nminus. \Mplus drug combinations have
lower frequencies (on average 5.520 as in Table~\ref{table:Statis_whole}), and higher and mostly
positive prediction values, while \Nminus drug combinations have higher frequencies (on average
42.082), and lower and mostly negative prediction values. For \Nminus, the mis-classification
typically happens on lower-frequency drug combinations (the mis-classification for \Mplus
does not show strong patterns with respect to drug combination frequencies).
As for \Mplus and \Nminus, drug combination frequencies are used to define 
\adr labels. 
Figure~\ref{fig:freq_pred_AD} shows that \GMopcm together are able to learn
and make predictions that correspond to drug combination frequencies and thus \adr labels.

Figure~\ref{fig:or_pred_BC} presents the \GMopcm prediction values with respect
to \OR values for \Mzero and \Nzero. \Mzero drug combinations have higher \OR values and
also higher and mostly positive prediction values, while \Nzero drug combinations have
lower \OR values and also lower and mostly negative prediction values. For \Nzero,
the mis-classification typically happens on drug combinations of higher \OR values (close to
1 and thus more lean toward \adr; the mis-classification for \Mzero does not show
strong patterns with respect to \OR values). As we use \OR to define \adr labels
on \Mzero and \Nzero, Figure~\ref{fig:or_pred_BC} shows \GMopcm is able to make
reasonably accurate prediction values on the drug combinations.

\subsection*{\Dscider drug enrichment}
%
%

%
Table~\ref{table:enrich} presents the average percentage of \Dscider drugs
among all the drug combinations.
For each drug combination, the percentage is calculated as 
the number of its drugs that can cause myopathy on their own (i.e., drugs in \Dscider)
divided by the drug combination order. 
As Table~\ref{table:enrich} shows, top-10 mis-classified
\Nneg drug combinations (i.e., \Nmist) have almost twice as many \Dscider drugs (30.7\%)
as those in \Nneg (15.6\%), and even more than those in \Mpos
drug combinations (24.3\%). 
In addition, mis-classified \Nneg drug combinations (i.e., \Nmis) also have 
significantly more \Dscider drugs (18.6\%) than those in \Nneg (15.6\%).
Since \GMopcm matches similar drugs, high \Dscider drug enrichment could be a
primary reason for the mis-classification. 

\subsection*{Top predictions}

%

\subsubsection*{Top mis-classification on \Nneg}
%

%
Table~\ref{table:top_com_d6k_miss} lists the top-10 (in terms of prediction values)
drug combinations in \Nneg (i.e., without
myopathy) that are mis-classified as positive (i.e., with myopathy) by \GMopcm.
For those drug combinations which appear in \Nzero,
we present their \OR values, otherwise only frequencies.
Those top mis-classified \Nneg drug combinations contain many single
drugs, which on their own can induce myopathy (i.e., in \Dscider,
bold in Table~\ref{table:top_com_d6k_miss}).
As a matter of fact, the percentage of \Dscider drugs in top mis-classified \Nneg
drug combinations is significantly higher than average. 
%
%
In Table~\ref{table:top_com_d6k_miss}, 
one special mis-classified \Nneg drug combination is
\{lansoprazole omeprazole pantoprazole rabeprazole\}, which does not contain any \Dscider
drugs. This set of drugs
is commonly used as proton pump inhibitors (PPIs)
to decrease the amount of acid produced in the stomach. 
Some case studies show evidence of causality between 
the PPI drug class and myopathy~\cite{Colmenares2017,Clark2006}.


\subsubsection*{Top prediction on \Mpos}

%
Table~\ref{table:top_com_d6k} presents the top-10 correctly predicted \Mpos
drug combinations by \GMopcm.
These drug combinations are significantly enriched with
\Dscider drugs (i.e., drugs that can induce myopathy on their own). As
Table~\ref{table:enrich} shows, \Mt has the most \Dscider drugs (89.8\%)
compared to all the other sets and significantly more than \Mpos. 
In particular, all of these combinations contain statin drugs
(e.g., atorvastatin, simvastatin and rosuvastatin, etc.).
These statin-related drugs have been studied in literature as a
drug class that has high possibilities to induce myopathy~\cite{Joy2009,Fernandez2011}.
In addition, in Table~\ref{table:top_com_d6k},
4 out of the 6 \Mzero drug combinations among top 10 (i.e., the drug combinations
that have \OR values) have their \OR values higher than average in \Mzero (31.998 as in
Table~\ref{table:Statis_whole}), and 3 out of the 4 \Mplus drug combinations among top 10
(i.e., the drug combinations that do not have \OR values) have their frequency higher
than average in \Mplus (5.520 as in Table~\ref{table:Statis_whole}).
In addition, among the top-20 drug combinations predicted by \GMopcm, 
7 out of 12 \Mzero drug
%
combinations have their \OR values higher than average in \Mzero, 
and 5 out of 8 \Mplus drug combinations have their frequency higher than average in \Mplus.
The average \OR values of the top-10, top-20 and top-50 drug combinations from \Mzero predicted 
by \GMopcm are 55.725, 42.956 and 42.114, respectively, 
and they are all higher than the average 31.998 for \Mzero.
The average frequencies of the top-10, top-20 and top-50 drug combinations from \Mplus
predicted by \GMopcm are 8.250, 6.875 and 6.524, respectively, and 
they are also all higher than the average 5.520 on \Mplus.
This indicates that \GMopcm does learn signals from \Mpos and correspondingly makes
predictions.

\subsubsection*{Top non-\Dscider prediction on \Mpos}
%
%
Table~\ref{table:top_com_d6k_without_M} presents the top-10 correctly predicted
\Mpos drug combinations by \GMopcm that do not contain any drugs from
\Dscider (i.e., do not contain drugs that can induce myopathy on their own).
4 out of 7 drug combinations from \Mzero in this table have their \OR values higher than the
average in \Mzero (31.998 as in Table~\ref{table:Statis_whole}). The 3 drug
combinations from \Mplus in this table have their frequency lower than the average in
\Mplus (5.520 as in Table~\ref{table:Statis_whole}) but very close.
In Table~\ref{table:top_com_d6k_without_M}, 8 out of top-10 drug combinations include alendronate.
Case studies demonstrate that several events of severe muscle pain, 
which is the common symptom of myopathy, were reported 
after patients started therapy with alendronate~\cite{Wysowski2005}, 
showing the association between the medical treatment with alendronate and myopathy.
%

\section*{Discussions}

The experimental results show that the new methods with drug co-medication based single drug similarities 
outperform other kernels, such as convolutional kernels ~\cite{Haussler1999} and probabilistic kernels ~\cite{Zhou2006},
and can accurately predict whether a drug combination is likely to induce ADRs of interest.
The experimental results demonstrate the advance of such single drug similarities that 
leverage co-medication patterns among
high-order drug-drug interactions, and also inspire further
exploration that learns such similarities in a pure data-driven fashion without
pre-defined kernels, for example, via manifold learning. Further research would also
include learning drug representations in a data-driven fashion such that the
representations better quantify drug similarities in terms of their co-medication
patterns. Deep learning would be an optimistic option for such drug representation
learning.

\section*{Conclusions}

In this manuscript,
SVM-based classification methods were developed to predict whether a drug combination of
arbitrary orders is likely to induce adverse drug reactions.
Novel kernels over drug combinations of arbitrary orders were developed for such
classification.
%
These kernels were constructed from various single-drug
information including drug co-medication patterns, and compare
drug combination similarities based on single drugs they
have and the relations among the single drugs.
Specifically, a novel kernel over drug combinations of arbitrary
orders was developed based on graph matching over drug combination graphs.
A dataset from FDA Adverse Event Reporting
System (FAERS) was constructed to test the new methods.
The experimental results demonstrated that the new methods
with drug co-medication based single drug similarities and
graph matching based kernels achieve the best AUC as 0.912.
The prediction also revealed strong patterns among drug combinations
(e.g., statin enriched) that may be highly correlated
with their induced ADRs.

\section*{List of abbreviations}
%
DDI: Drug-Drug Interactions; ADR: Adverse Drug Reaction; SDS: Single Drug Similarities;
ECFP: Extended Connectivity Fingerprints; and \OR: Odds Ratio.

\section*{Declarations}
%
\begin{backmatter}
%
\section*{Ethics approval and consent to participate}

	Not applicable
\section*{Consent for publication}
	Not applicable
\section*{Availability of data and material}
The data and materials will be made publicly available
upon the acceptance of the manuscript. 
\section*{Competing interests}
	The authors declare that they have no competing interests.
\section*{Funding}
  This material is based upon work supported by the National Science Foundation
  under Grant Number IIS-1566219 and
  IIS-1622526. Any opinions, findings, and conclusions or recommendations
  expressed in this material are those of the author(s) and do not necessarily reflect
  the views of the National Science Foundation.		 
%

\section*{Author's contributions}
Wen-Hao Chiang implemented the methods and conducted the experiments. Li Shen, Lang Li
and Xia Ning
developed the methods and designed the experiments. Xia Ning analyzed the experimental results.
Wen-Hao Chiang and Xia Ning wrote the manuscript.


\bibliographystyle{bmc-mathphys} 
\bibliography{paper}      




\clearpage

\section*{Figures}

\begin{figure*}[h!]
\mbox{
\scalebox{1.5}{ \hspace{-650pt} 


\usetikzlibrary{decorations.markings}

\usetikzlibrary[topaths]
\usetikzlibrary{calc}

\tikzstyle{vertex}=[    draw,
                        circle,
                        text=black,
                        minimum width=10pt]

\definecolor{c1}{RGB}{0,0,255}
\definecolor{c2}{RGB}{0,255,0}
\definecolor{c3}{RGB}{255,0,0}
\definecolor{c4}{RGB}{255, 255, 0}
\definecolor{c5}{RGB}{255, 255, 255}


  \small
 \begin{tikzpicture}[
	auto,node distance=3cm,
	scale = 0.6,
  	thick,main node/.style={circle,draw,font=\sffamily\Large\bfseries} ]

	\node[vertex, fill=c1]   (l1)  at (2.5,   0.5) {};
	\node[vertex, fill=c2]   (l2)  at (5.0,   2.0) {};
	\node[vertex, fill=c3]   (l3)  at (4.0,   4.5) {};
	\node[vertex, fill=c4]   (l4)  at (1.0,   4.5) {};	
	\node[vertex, fill=c5]   (l5)  at (0.0,   2.0) {};

	\node[]   (ll1)  at ( 1.6,   0.5) {$\drug_{p1}$};
	\node[]   (ll2)  at ( 5.9,   2.4) {$\drug_{p2}$};
	\node[]   (ll3)  at ( 4.9,   4.9) {$\drug_{p3}$};
	\node[]   (ll4)  at ( 0.1,   4.9) {$\drug_{p4}$};
	\node[]   (ll5)  at (-0.9,   2.2) {$\drug_{p5}$};

        \node (D1) at (2.5,   -0.5) {\large{\Gp (\Dp)}};

	\path (l1) edge (l2);
	\path (l1) edge (l3);
	\path (l1) edge (l4);
        \path (l1) edge (l5);
        \path (l2) edge (l3);
        \path (l2) edge (l4);
        \path (l2) edge (l5);
        \path (l3) edge (l4);
        \path (l3) edge (l5);
        \path (l4) edge (l5);

        \node[vertex, fill=c1!20]   (r1)  at (8.5,   0.5)  {};
        \node[vertex, fill=c2!20]   (r2)  at (11.0,   2.0) {};
        \node[vertex, fill=c3!20]   (r3)  at (10.0,   4.5) {};

        \node[]   (rr1)  at ( 9.4,   0.5) {$\drug_{q1}$};
        \node[]   (rr2)  at (11.9,   2.0) {$\drug_{q2}$};
        \node[]   (rr3)  at (10.9,   4.5) {$\drug_{q3}$};

	\node (D2) at (9.75,   -0.5) {\large{\Gq (\Dq)}};	

	\path (r1) edge (r2);
	\path (r1) edge (r3);
	\path (r2) edge (r3);

	\foreach \numbera in {1,...,3}{
		\path [dotted] (l\numbera) edge (r\numbera);
	}

  \end{tikzpicture}


}
}
\caption{\csentence{Graph matching for drug combinations}}
\label{fig:gmatch}
\end{figure*}
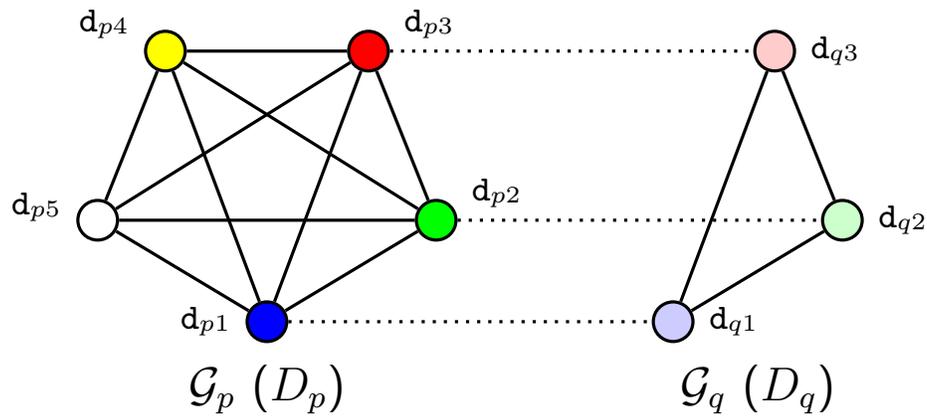

\begin{figure}[h!]
\input{./figure/K_cosim_D6k/Order_A_D/K_cosim_D6k_Order_vs_Prediction_value_A_D.tex}
\caption{\csentence{Orders vs. predictions in \GMopcm}}
\label{fig:order_pred_AD}
\end{figure}

\begin{figure}[h!]
\input{./figure/K_cosim_D6k/Order_B_C/K_cosim_D6k_Order_vs_Prediction_value_B_C.tex}
\caption{\csentence{Orders vs. predictions in \GMopcm}}
\label{fig:order_pred_BC}
\end{figure}

\begin{figure}[h!]
\input{./figure/K_cosim_D6k/Freq_A_D/K_cosim_D6k_Freq_vs_Prediction_value_A_D.tex}
\caption{\csentence{Frequencies vs. predictions in \GMopcm}}
\label{fig:freq_pred_AD}
\end{figure}

\begin{figure}[h!]
\input{./figure/K_cosim_D6k/OR_B_C/K_cosim_D6k_OR_vs_Prediction_value_B_C.tex}
\caption{\csentence{\OR values vs. predictions in \GMopcm}}
\label{fig:or_pred_BC}
\end{figure}


\section*{Additional Files}
  \subsection*{Additional file 1 --- Drug-Drug Interaction Prediction based on Co-Medication Patterns and Graph Matching (Supplementary Materials)}
	The additional file named as ``supp.pdf'' includes more experimental results on other datasets and it is provided in PDF format .
	Any PDF reader are recommended to view the file.




        \section*{Tables}
        \begin{table}[ht!]
  \caption{Table of Notations}
  \label{tbl:Notation}
  \centering
  \begin{threeparttable}
  \renewcommand\arraystretch{1.2}
    \begin{small}
      \begin{tabular}{
	@{\hspace{2pt}}l@{\hspace{10pt}}
	@{\hspace{2pt}}p{0.79\linewidth}@{\hspace{ 2pt}}
	}
     \hline\hline
	    {Notation}					& Description	\\
		\hline
		\drug					& Drug\\
		\D					& Drug combination\\
		$\mathcal{G}$				& Complete graph for a drug combination\\		
		$\SDSd$					& Single drug similarity from drug 2d structures\\
		$\SDSc$                                 & Single drug similarity based on co-medications\\
		\GMop					& Kernel based on graph matching algorithm\\
		\GMtan					& Kernel from common drugs\\
		\GMavg					& Kernel from drug similarities\\
		\GMprob					& Probabilistic drug combination kernel\\
    \hline\hline
      \end{tabular}
      \begin{tablenotes}
        \setlength\labelsep{0pt}
	\begin{footnotesize}
	\item
		  ~~
          \par
	\end{footnotesize}
      \end{tablenotes}
    \end{small}
  \end{threeparttable}
\end{table}

\begin{table}[ht!]
  \caption{Contingency Table}
  \label{table:contigen}
  \centering
  \begin{threeparttable}
    \begin{small}
      \begin{tabular}{
	@{\hspace{2pt}}c@{\hspace{7pt}}|
	@{\hspace{2pt}}c@{\hspace{8pt}}          
	@{\hspace{2pt}}c@{\hspace{8pt}}|
	@{\hspace{2pt}}c@{\hspace{8pt}}
	}
        \hline\hline
        $\OR = \frac{n_1}{m_1}/ \frac{n_2}{m_2}$	& \adr			& no \adr		& total	\\
        \hline
        \D					& $n_{1}$ 		& $m_{1}$		& $n_{1}+m_{1}$\\
         $\setminus$\D 				& $n_{2}$     		& $m_{2}$ 		& $n_{2}+m_{2}$\\
	\hline
	total					& $n_{1} + n_{2}$	& $m_{1} + m_{2}$	& $n_{1}+n_{2}+m_{1}+m_{2}$\\
        \hline\hline
      \end{tabular}
      \begin{tablenotes}
        \setlength\labelsep{0pt}
	\begin{footnotesize}
	\item
	  In the table, $n_{1}$ is the number of events where \D is taken with \adr occurring; 
	$m_{1}$ is the number of events where \D is taken without \adr occurring;
	$n_{2}$ is the number of events where \D is not taken with \adr occurring;
	  and $m_{2}$ is the number of events where \D is not taken without \adr occurring, respectively.
          \par
	\end{footnotesize}
      \end{tablenotes}
    \end{small}
  \end{threeparttable}
\end{table}

\begin{table}[ht!]
\vspace{0mm}
  \caption{Data Statistics}
  \label{table:Statis_whole}
  \centering
  \begin{threeparttable}
    \begin{small}
      \begin{tabular}{
	@{\hspace{5pt}}l@{\hspace{6pt}} 
	@{\hspace{5pt}}c@{\hspace{6pt}}
	@{\hspace{5pt}}r@{\hspace{6pt}}          
	@{\hspace{5pt}}r@{\hspace{6pt}}
        @{\hspace{2pt}}r@{\hspace{2pt}}
	@{\hspace{5pt}}r@{\hspace{6pt}}
	@{\hspace{5pt}}r@{\hspace{6pt}}
	}
        \hline\hline
        \multirow{2}{*}{dataset} & \multirow{2}{*}{stats} & \multicolumn{2}{c}{\Nneg} && \multicolumn{2}{c}{\Mpos} \\
        \cline{3-4} \cline{6-7}
	              & & \Nminus	& \Nzero	&& \Mzero        & \Mplus \\
        \hline
	&\#$\{\D\}$     & 621,449	& 1,264		&& 8,986		& 27,387		\\
 	&\#$\{\drug\}$  & 1,209         & 417           && 881           & 1,201                 \\
\Dall	&avgOrd	& 6.100		& 2.351     	&& 3.588         & 7.096		\\
	&avgFrq	& 1.761		& 225.317      	&& 13.730        & 1.402		\\
	&avg\OR		& -	       	& 0.546		&& 16.343        & -	 		\\
	\cline{2-6}
        \hline
        &\#$\{\D\}$	& 2,200         & 1,264		&& 2,464 	 & 1,000                 \\
        &\#$\{\drug\}$	& 562           & 417           &&  692           &   679                 \\
\Dataset   &avgOrd    & 2.678         & 2.351         && 3.809          & 7.615     \\
        &avgFrq        & 42.082        & 225.317       && 20.565         & 5.520        \\
        &avg\OR         & -             & 0.546         &&  31.998        & -        \\

        \hline\hline
      \end{tabular}
      \begin{tablenotes}
        \setlength\labelsep{0pt}
	\begin{footnotesize}
	\item         
	  In this table, ``\#$\{\D\}$'' and ``\#$\{\drug\}$'' represent the number of drug combinations and the number of involved drugs, respectively.
	  In each set of drug combinations, ``avgOrd'' is the the average order; ``avgFrq'' is the average frequency; and 
	  ``avgOR'' represents the average \OR. 
          \par
	\end{footnotesize}
      \end{tablenotes}
    \end{small}
  \end{threeparttable}
\vspace{-0mm}
\end{table}

\begin{table}[ht!]
\vspace{-0mm}
  \centering
  \caption{Overall Performance Comparison}
  \label{table:result_6k}
  \begin{threeparttable}
    \begin{small}
      \begin{tabular}[]{
          @{\hspace{3pt}}l@{\hspace{3pt}}
	  @{\hspace{0pt}}c@{\hspace{0pt}}
          @{\hspace{1pt}}c@{\hspace{3pt}}
          @{\hspace{1pt}}c@{\hspace{1pt}}
          @{\hspace{1pt}}c@{\hspace{3pt}}
          @{\hspace{1pt}}c@{\hspace{1pt}}
          @{\hspace{1pt}}c@{\hspace{3pt}}
          @{\hspace{1pt}}c@{\hspace{1pt}}
          @{\hspace{1pt}}c@{\hspace{3pt}}
          @{\hspace{1pt}}c@{\hspace{1pt}}
          @{\hspace{1pt}}c@{\hspace{3pt}}
          @{\hspace{1pt}}c@{\hspace{1pt}}
          @{\hspace{1pt}}c@{\hspace{3pt}}
        }
        \hline\hline
	\multicolumn{1}{c}{$\mathcal{K}$} && \multicolumn{2}{c}{\GMop}   && \multicolumn{2}{c}{\GMtan}   && \multicolumn{2}{c}{\GMavg}   && \multicolumn{2}{c}{\GMprob} \\
 	\cline{1-1} \cline{3-4}                 \cline{6-7}                 \cline{9-10}                  \cline{12-13}
        ~~\SDS && \IID & \coSim && \orderI & \orderII && \IID & \coSim && \IID & \coSim \\

	\hline

acc  && 0.829 & \textbf{0.836} && 0.817 & 0.827 && 0.827          & 0.825          && 0.763 & 0.765  \\
pre  && 0.889 & 0.892          && 0.879 & 0.878 && \textbf{0.893} & 0.865          && 0.810 & 0.770  \\
rec  && 0.752 & 0.765          && 0.735 & 0.759 && 0.744          & \textbf{0.770} && 0.689 & 0.756  \\
F1   && 0.815 & \textbf{0.823} && 0.801 & 0.814 && 0.812          & 0.815          && 0.744 & 0.763  \\
AUC  && 0.898 & \textbf{0.912} && 0.907 & 0.909 && 0.900          & 0.900          && 0.843 & 0.853  \\

        \hline
        \hline
      \end{tabular}
      \begin{tablenotes}
        \setlength\labelsep{0pt}
	\begin{footnotesize}
	\item
	In this table, ``acc'', ``pre'', ``rec'', ``F1'' and ``AUC'' represent 
	accuracy, precision, recall, F1 and the area under an receiver operating characteristic curve, respectively.
        \GMtan with ``\orderI'' corresponds to \GMtanorderI, and
        \GMtan with ``\orderII'' corresponds to \GMtanorderII. 
          \par
	\end{footnotesize}
      \end{tablenotes}
    \end{small}
  \end{threeparttable}
\end{table}

\begin{table}[ht!]
  \centering
  \caption{Average Percentage (\%) of \Dscider Drugs (\GMopcm)}
  \label{table:enrich}
    \bgroup
    \def\arraystretch{1.6}

  \begin{threeparttable}
    \begin{small}
      \begin{tabular}{
	@{\hspace{0pt}}c@{\hspace{5pt}}
	@{\hspace{5.6pt}}c@{\hspace{5pt}}   
	@{\hspace{5.6pt}}c@{\hspace{5pt}}       
	@{\hspace{5.6pt}}c@{\hspace{5pt}}
	@{\hspace{5.6pt}}c@{\hspace{5pt}}
	@{\hspace{5.6pt}}c@{\hspace{5pt}}
        @{\hspace{5.6pt}}c@{\hspace{5pt}}
	@{\hspace{5.6pt}}c@{\hspace{0pt}}
	}
        \hline\hline
		\Mmist	& \Mmis		 & \Mpos 	&\Mt   & \Nmist  &\Nmis & \Nneg &\Nt \\
        \hline
	   13.3 & 16.6 &24.3 & 89.8 & 30.7 & 18.6 & 15.6 & 0.10 \\
	\hline\hline

      \end{tabular}
      \begin{tablenotes}
        \setlength\labelsep{0pt}
	\begin{footnotesize}
	\item 
	  \Mmis and \Nmis are the sets of all mis-classified drug combinations in \Mpos and \Nneg
          by \GMopcm, respectively.
	  \Mmist is the set of the top-10 mis-classified drug combinations in \Mmis and 
	  \Nmist is the set of the top-10 mis-classified drug combinations in \Nmis by \GMopcm.
	  \Mt and \Nt are the sets of correctly classified drug combinations in \Mpos and \Nneg with the top-10 highest and lowest prediction values, respectively, by \GMopcm.	  	
          \par
	\end{footnotesize}
      \end{tablenotes}
    \end{small}
  \end{threeparttable}
    \egroup
\vspace{-0mm}
\end{table}

\begin{table}[h!]
  \centering
  \caption{Top mis-Classified \Nneg Drug Combinations by \GMopcm}
  \label{table:top_com_d6k_miss}
  \begin{threeparttable}
    \begin{small}
      \begin{tabular}{
 	@{\hspace{0pt}}r@{\hspace{3pt}}
        @{\hspace{3pt}}r@{\hspace{3pt}}
        @{\hspace{3pt}}r@{\hspace{3pt}}
	@{\hspace{3pt}}r@{\hspace{3pt}}
        @{\hspace{3pt}}p{0.64\linewidth}@{\hspace{0pt}}
	}
        \hline\hline
N       & prd &frq &\OR     & combinations          \\
        \hline
        1&2.696 & 26 & -       &\textbf{atorvastatin} \textbf{fenofibrate} rosiglitazone \textbf{simvastatin}  \\
        2&2.507 & 26 & -       &\textbf{allopurinol} amlodipine \textbf{atorvastatin} levothyroxine naproxen omeprazole \textbf{simvastatin}  \\
        3&1.878 & 22 & -       &acetylsalicylicacid \textbf{atorvastatin} bisoprolol clopidogrel ramipril \textbf{simvastatin}  \\
        4&1.855 & 27 & -       &acetylsalicylicacid atenolol \textbf{atorvastatin} furosemide lansoprazole lisinopril nitroglycerin  \\
        5&1.785 & 21 & -       &\textbf{citalopram} clozapine isosorbidemononitrate prochlorperazine \textbf{simvastatin} \textbf{zopiclone}  \\
        6&1.750 & - & 0.842    &amlodipine bisoprolol \textbf{pravastatin} ramipril \textbf{simvastatin} spironolactone warfarin  \\
        7&1.696 & 22 & -       &amlodipine clopidogrel ibuprofen omeprazole ramipril \textbf{simvastatin}  \\
        8&1.669 & 29 & -       &bisoprolol flecainide ramipril \textbf{simvastatin}  \\
        9&1.613 & 35 & -       &\textbf{aripiprazole} \textbf{atorvastatin} bendroflumethiazide clozapine diazepam folicacid furosemide iron lactulose lansoprazole perindopril ramipril trimethoprim \textbf{zopiclone}  \\
        10&1.549& - & 0.875    &lansoprazole omeprazole pantoprazole rabeprazole  \\
        \hline\hline
      \end{tabular}
      \begin{tablenotes}
        \setlength\labelsep{0pt}
	\begin{footnotesize}
	\item
	 In this table,  ``prd'', ``frq'' and ``\OR'' represent prediction values, frequency and odds ratio, respectively.
	Drugs in \Dscider are marked in \textbf{bold}.  
          \par
	\end{footnotesize}
      \end{tablenotes}
    \end{small}
  \end{threeparttable}
\end{table}

\begin{table}[h!]
  \centering
  \caption{Top Predictions on \Mpos by \GMopcm}
  \label{table:top_com_d6k}
  \begin{threeparttable}
    \begin{small}
      \begin{tabular}{
        @{\hspace{0pt}}r@{\hspace{3pt}}
        @{\hspace{3pt}}r@{\hspace{3pt}}
        @{\hspace{3pt}}r@{\hspace{3pt}}
	@{\hspace{3pt}}r@{\hspace{3pt}}
        @{\hspace{3pt}}p{0.6\linewidth}@{\hspace{0pt}}
	}
        \hline\hline
	N       & prd   & frq &\OR & Combinations                \\
        \hline
1 &4.167 & 3  &  -      &\textbf{atorvastatin} lansoprazole \textbf{pravastatin} \textbf{rosuvastatin} \textbf{simvastatin}  \\
2 &4.009 & -  & 11.372  &\textbf{atorvastatin} \textbf{pravastatin} \textbf{rosuvastatin} \textbf{simvastatin}  \\
3 &3.776 & -  & 50.043  &\textbf{atorvastatin} \textbf{fenofibrate} metformin \textbf{pravastatin} \textbf{rosuvastatin} \textbf{simvastatin}  \\
4 &3.734 & -  & 68.232  &\textbf{atorvastatin} metformin \textbf{pravastatin} \textbf{rosuvastatin} \textbf{simvastatin}  \\
5 &3.676 & -  & 45.487  &\textbf{atorvastatin} \textbf{lovastatin} \textbf{rosuvastatin} \textbf{simvastatin}  \\
6 &3.618 & -  & 136.470 &\textbf{atorvastatin} \textbf{pravastatin} \textbf{rosuvastatin} \textbf{simvastatin} tadalafil  \\
7 &3.573 & 9  &  -      &\textbf{atorvastatin} \textbf{fenofibrate} \textbf{pravastatin} \textbf{simvastatin}  \\
8 &3.552 & 10 &  -      &\textbf{atorvastatin} \textbf{ezetimibe} \textbf{fenofibrate} \textbf{rosuvastatin}  \\
9 &3.519 & -  & 22.746  &\textbf{atorvastatin} \textbf{ezetimibe} \textbf{rosuvastatin} \textbf{simvastatin}  \\
10&3.461 & 11 &  -      &\textbf{atorvastatin} lansoprazole \textbf{pravastatin} \textbf{simvastatin}  \\

        \hline\hline
      \end{tabular}
      \begin{tablenotes}
        \setlength\labelsep{0pt}
	\begin{footnotesize}
	\item
          In this table, ``prd'', ``frq'' and ``\OR'' represent prediction values, frequency and odds ratio, respectively.
          Drugs in \Dscider are marked in \textbf{bold}.

          \par
	\end{footnotesize}
      \end{tablenotes}
    \end{small}
  \end{threeparttable}
\vspace{-0mm}
\end{table}

\begin{table}[h!]
  \centering
  \caption{Top Predictions without \Dscider Drugs by \GMopcm}
\setlength{\abovecaptionskip}{-100pt} 
  \label{table:top_com_d6k_without_M}
  \begin{threeparttable}
    \begin{small}
      \begin{tabular}{
        @{\hspace{0pt}}r@{\hspace{3pt}}
        @{\hspace{3pt}}r@{\hspace{3pt}}
        @{\hspace{3pt}}r@{\hspace{3pt}}
        @{\hspace{3pt}}r@{\hspace{3pt}}
        @{\hspace{3pt}}p{0.60\linewidth}@{\hspace{0pt}}
	}
        \hline\hline
N       & prd   & frq &\OR & Combinations                \\
        \hline
1 &2.083  &4 &- &calcium clonazepam colestipol prednisone teriparatide  \\
2 &1.992  &- &45.487  &alendronate anastrozole desloratadine hydrochlorothiazide lisinopril triamterene valdecoxib vitaminc  \\
3 &1.968  &- &17.058  &alendronate raloxifene risedronate teriparatide  \\
4 &1.960  &- &90.978  &alendronate amlodipine atenolol clonazepam raloxifene teriparatide  \\
5 &1.901  &- &45.489  &alendronate fexofenadine hydrochlorothiazide omeprazole prednisone risedronate triamterene  \\
6 &1.850  &5 &-       &alendronate fexofenadine levothyroxine nabumetone oxybutynin  \\
7 &1.849  &- &113.720 &alendronate calcium esomeprazole ibandronate levothyroxine rabeprazole  \\
8 &1.843  &- &7.581  &alendronate calciumgluconate teriparatide  \\
9 &1.838  &- &22.744 &alendronate calcium levothyroxine raloxifene teriparatide  \\
10&1.834  &4 &- &calcium escitalopram iron ketorolac raloxifene teriparatide  \\

        \hline\hline
      \end{tabular}
      \begin{tablenotes}
        \setlength\labelsep{0pt}
	\begin{footnotesize}
	\item
	In this table, ``prd'', ``frq'' and ``\OR'' represent prediction values, frequency and odds ratio, respectively.
          \par
	\end{footnotesize}
      \end{tablenotes}
    \end{small}
  \end{threeparttable}
\vspace{-0mm}
\end{table}

\end{backmatter}
\end{document}


\begin{frontmatter}

\begin{fmbox}
\dochead{Research}


\title{\mytitle}


\author[
   addressref={aff1},                   
   email={chiangwe@iupui.edu}   
]{\inits{WC}\fnm{Wen-Hao} \snm{Chiang}}
\author[
   addressref={aff2},
   email={Li.Shen@pennmedicine.upenn.edu}
]{\inits{LS}\fnm{Li} \snm{Shen}}
\author[
   addressref={aff3},
   email={Lang.Li@osumc.edu}
]{\inits{LS}\fnm{Lang} \snm{Li}}
\author[
   addressref={aff4},
   corref={aff4},
   email={xning@iupui.edu}
]{\inits{XN}\fnm{Xia} \snm{Ning}}


\address[id=aff1]{
  \orgname{Department of Computer \& Information Science, Indiana University - Purdue University Indianapolis}, 
  \postcode{46202}                                
  \city{Indianapolis},                              
  \cny{USA. Email: chiangwe@iupui.edu}                                    
}
\address[id=aff2]{%
  \orgname{Department of Biostatistics, Epidemiology and Informatics, University of Pennsylvania},
  \postcode{19104}
  \city{Philadelphia},
  \cny{USA. Email: Li.Shen@pennmedicine.upenn.edu}
}
\address[id=aff3]{%
  \orgname{Department of Biomedical Informatics, Ohio State University},
  \postcode{43210}
  \city{Columbus},
  \cny{USA. Email: Lang.Li@osumc.edu}
}
\address[id=aff4]{
  \orgname{Department of Computer \& Information Science, Indiana University - Purdue University Indianapolis}, 
  \postcode{46202}                                
  \city{Indianapolis},                              
  \cny{USA. Email: xning@iupui.edu}                                    
}


\begin{artnotes}
\end{artnotes}

\end{fmbox}

\end{frontmatter}
\section*{Performance on Other Datasets}

\begin{table}[!h]
  \caption{Data Statistics of \Dtwok and \Dfourk}
  \label{table:Statis_2k_4k}
  \centering
  \begin{threeparttable}
    \begin{small}
      \begin{tabular}{
	@{\hspace{5pt}}l@{\hspace{5pt}} 
	@{\hspace{5pt}}c@{\hspace{5pt}}
	@{\hspace{5pt}}r@{\hspace{5pt}}          
	@{\hspace{5pt}}r@{\hspace{5pt}}
        @{\hspace{2pt}}r@{\hspace{2pt}}
	@{\hspace{5pt}}r@{\hspace{5pt}}
	@{\hspace{5pt}}r@{\hspace{5pt}}
	}
        \hline\hline
        \multirow{2}{*}{dataset} & \multirow{2}{*}{stats} & \multicolumn{2}{c}{\Nneg} && \multicolumn{2}{c}{\Mpos} \\
        \cline{3-4} \cline{6-7}
	              & & \Nminus	& \Nzero	&& \Mzero        & \Mplus \\
        \hline
	&\#$\{\D\}$     & 1,000		& 1,000		&& 1,000      & 1,000	\\
 	&\#$\{\drug\}$  & 1,000  	& 369   	&& 596      &   679    \\
\Dfourk	&avgOrd		&  2.585       	& 2.340     	&& 4.770     & 7.615	\\
	&avgFrq		& 62.545       	& 264.898     	&& 5.957    & 5.520	\\
	&avg\OR		& -     	& 0.451		&& 60.994    & -		\\
	\cline{2-6}
        \hline
        &\#$\{\D\}$	& 1,000   & -       && -	      & 1,000                 \\
        &\#$\{\drug\}$	&   426   & -       && -	      &   679                 \\
\Dtwok  &avgOrd      	& 2.585   & -       && -      	      & 7.615      \\
        &avgFrq         & 62.545  & -       && -	      & 5.520        \\
        &avg\OR         & -       & -       && -	      & -   \\

        \hline\hline
      \end{tabular}
      \begin{tablenotes}
        \setlength\labelsep{0pt}
	\begin{footnotesize}
	\item         
	  In this table, ``\#$\{\D\}$'' and ``\#$\{\drug\}$'' represent the number of drug combinations and the number of involved drugs, respectively.
	  In each set of drug combinations, ``avgOrd'' is the the average order; ``avgFrq'' is the average frequency; and 
	  ``avgOR'' represents the average \OR. 
          \par
	\end{footnotesize}
      \end{tablenotes}
    \end{small}
  \end{threeparttable}
\end{table}
\begin{table}[t]
  \centering
  \caption{Overall Performance Comparison of \Dtwok}
  \label{table:result_2k_and_4k}
  \begin{threeparttable}
    \begin{small}
      \begin{tabular}[]{
          @{\hspace{0pt}}l@{\hspace{1pt}}
	  @{\hspace{0pt}}l@{\hspace{1pt}}
          @{\hspace{1pt}}c@{\hspace{1pt}}
          @{\hspace{1pt}}c@{\hspace{1pt}}
          @{\hspace{1pt}}c@{\hspace{1pt}}
          @{\hspace{1pt}}c@{\hspace{1pt}}
          @{\hspace{1pt}}c@{\hspace{1pt}}
          @{\hspace{1pt}}c@{\hspace{1pt}}
          @{\hspace{1pt}}c@{\hspace{1pt}}
          @{\hspace{1pt}}c@{\hspace{1pt}}
          @{\hspace{1pt}}c@{\hspace{1pt}}
          @{\hspace{1pt}}c@{\hspace{1pt}}
          @{\hspace{1pt}}c@{\hspace{1pt}}
          @{\hspace{1pt}}c@{\hspace{0pt}}
        }
        \hline\hline
      	   \multicolumn{2}{c}{$\mathcal{K}$} && \multicolumn{2}{c}{\GMop}   && \multicolumn{2}{c}{\GMtan}   && \multicolumn{2}{c}{\GMavg}   && \multicolumn{2}{c}{\GMprob} \\
        \cline{1-2} \cline{4-5}                 \cline{7-8}                 \cline{10-11}                  \cline{13-14}
         \multicolumn{2}{c}{$\SDS$} && \IID & \coSim && \orderI & \orderII && \IID & \coSim && \IID & \coSim \\

        \hline

	&acc  &&  0.929 & \textbf{0.938}&& 0.897 & 0.910 && 0.924 & 0.912 && 0.845 & 0.860  \\
	&pre  &&  0.954 & \textbf{0.959}&& 0.941 & 0.942 && 0.957 & 0.917 && 0.839 & 0.860  \\
\Dtwok  &rec  &&  0.902 & \textbf{0.916}&& 0.847 & 0.873 && 0.888 & 0.907 && 0.854 & 0.869  \\
	&F1   &&  0.927 & \textbf{0.937}&& 0.891 & 0.906 && 0.921 & 0.912 && 0.846 & 0.861  \\
	&AUC  &&  0.973 & \textbf{0.976}&& 0.964 & 0.968 && 0.970 & 0.960 && 0.920 & 0.926  \\
	\hline
	&acc  &&  0.896          &  \textbf{0.899} &&  0.867 &  0.882 &&  0.887           &  0.880  &&  0.817 &  0.821 \\
	&pre  &&  0.933          &  0.939          &&  0.928 &  0.930 &&  \textbf{0.944}  &  0.903  &&  0.838 &  0.816 \\
\Dfourk	&rec  &&  0.853          &  \textbf{0.854} &&  0.795 &  0.826 &&  0.822           &  0.851  &&  0.787 &  0.829 \\
	&F1   &&  0.891          &  \textbf{0.894} &&  0.856 &  0.875 &&  0.879           &  0.876  &&  0.812 &  0.822 \\
	&AUC  &&  0.953          &  \textbf{0.962} &&  0.954 &  0.958 &&  0.951           &  0.945  &&  0.898 & 0.908  \\

	\hline
        \hline
      \end{tabular}
      \begin{tablenotes}
        \setlength\labelsep{0pt}
	\begin{footnotesize}
	\item
In this table, ``acc'', ``pre'', ``rec'' and ``AUC'' represent
        accuracy, precision, recall, and the area under an receiver operating characteristic curve, respectively.
        In \GMtan, ``\orderI'' is the kernel, in which the similarity between two drug combinations is calculated by the feature vectors, whose dimensions correspond to only one drug.
        As for ``\orderII'', the similarity between two drug combinations is calculated by the feature vectors, which have dimensions that correspond to different pairs of drugs.

          \par
	\end{footnotesize}
      \end{tablenotes}
    \end{small}
  \end{threeparttable}
\vspace{-2mm}
\end{table}
%


In \Dataset, we retain the 1,000 drug combinations from \Mplus and 
the 1,000 drug combinations with the largest \OR values from \Mzero. 
The labels of these drug combinations from \Mplus and \Mzero remain positive.
From \Nminus in \Dataset, we retain the top 1,000 most frequent drug combinations.
From \Nzero in \Dataset, we retain the 1,000 drug combinations with the smallest \OR values.
The labels of these drug combinations from \Nminus and \Nzero remain negative.
The pruned dataset is denoted as \Dfourk.

In \Dfourk, we retain the 1,000 drug combinations from \Mplus and 1,000 drug combinations from \Nminus and 
the labels remain positive and negative, respectively.
This pruned dataset based on \Dfourk is denoted as \Dtwok.
Table~\ref{table:Statis_2k_4k} presents the statistics of \Dfourk and \Dtwok.

Table~\ref{table:result_2k_and_4k} presents the overall performance comparison on dataset 
\Dtwok and \Dfourk.
In both dataset, kernel \GMop with \SDSc outperforms others in accuracy, recall, F1 and AUC.
Specifically, in accuracy, \GMop with \SDSc outperforms the second best kernel \GMop with \SDSd
at 0.97\% and 0.33\% in \Dtwok and \Dfourk, respectively.
In recall, \GMop with \SDSc outperforms the second best kernel \GMop with \SDSd 
at 1.55\% and 0.12\% in \Dtwok and \Dfourk, respectively.
In F1, \GMop with \SDSc outperforms the second best kernel \GMop with \SDSd
at 1.08\% and 0.34\% in \Dtwok and \Dfourk, respectively.
In AUC, \GMop with \SDSc outperforms the second best kernel \GMop with \SDSd
at 0.31\% in \Dtwok and outperforms the second best kernel \orderII \GMtan at 0.42\% in \Dfourk.
In precision, the best kernel \GMop with \SDSc outperforms the second best kernel \GMavg with \SDSd 
at 0.21\% in \Dtwok, whereas the best kernel \GMavg with \SDSd outperforms the second best kernel 
\GMop with \SDSc at 0.53\% in \Dfourk.
In three datasets, \Dtwok, \Dfourk and \Dataset, in general, \GMop with \SDSc has the best performance compared to other kernels
with a few exceptions. 
This may show the effectiveness to classify drug combinations by measuring the similarities between graphs, which 
represent drug combinations.

In kernel \GMop and \GMprob, \SDSc outperforms \SDSd on average. 
This indicates the consistency in three datasets of better capability to measure
the similarities based on co-medication patterns than drug 2D structures.
For kernel \GMtan, \orderII outperforms \orderI in both datasets as in \Dataset.
This consists with our observation before.
That is, the \orderII representation, which contains drug pairs as a feature,
can emphasize the co-occurrence patterns in drug combinations.
\bibliographystyle{bmc-mathphys} 
\renewcommand{\refname}{}
\bibliography{supp}